\documentclass[fleqn,10pt]{wlscirep}
\usepackage[utf8]{inputenc}
\usepackage[T1]{fontenc}
\usepackage{pdfpages}
\usepackage{multirow}
\usepackage{svg}
\usepackage{color}
\usepackage{subcaption}
\usepackage{smartdiagram}
\usepackage{tabularx}
\svgsetup{inkscapelatex=false}
\usepackage{lineno}

\newcommand{\myparagraph}[1]{\vspace*{1mm}\noindent {\bf #1} }

\title{Machine learning and Topological data analysis identify unique features of human papillae in 3D scans}

\newcolumntype{L}{>{\centering\arraybackslash}m{3cm}}

\author[1]{Rayna Andreeva}
\author[2]{Anwesha Sarkar}
\author[1,*]{Rik Sarkar}
\affil[1]{School of Informatics, University of Edinburgh, Edinburgh, United Kingdom}
\affil[2]{Food Colloids and Bioprocessing Group, School of Food Science and Nutrition, University of Leeds, Leeds, United Kingdom}

\affil[*]{rsarkar@inf.ed.ac.uk}

\begin{abstract}

The tongue surface houses a range of papillae that are integral to the mechanics and chemistry of taste and textural sensation. Although gustatory function of papillae is well investigated, the uniqueness of papillae within and across individuals remains elusive. Here, we present the first machine learning framework on 3D microscopic scans of human papillae ($n=2092$), uncovering the uniqueness of geometric and topological features of papillae. The finer differences in shapes of papillae are investigated computationally based on a number of features derived from discrete differential geometry and computational topology. 

Interpretable machine learning techniques show that persistent homology features of the papillae shape are the most effective in predicting the biological variables. Models trained on these features with small volumes of data samples predict the type of papillae with an accuracy of 85\%. The papillae type classification models can map the spatial arrangement of filiform and fungiform papillae on a surface. Remarkably, the papillae are found to be distinctive across individuals and an individual can be identified with an accuracy of 48\% among the 15 participants from a single papillae. Collectively, this is the first unprecedented evidence demonstrating that tongue papillae can serve as a unique identifier inspiring new research direction for food preferences and oral diagnostics.

\end{abstract}
\begin{document}

\flushbottom
\maketitle
%
%
\thispagestyle{empty}


\section*{Introduction}

The tongue is a highly sophisticated, heterogeneous anatomical structure and its operation is fundamental to speech, friction regulation and oral processing of food. The surface of the tongue is covered with tiny projections known as {\em papillae} which enable perception of taste, texture and oral mechanics. Of these numerous anatomical projections, {\em fungiform papillae} are considered as phenotypic markers of chemosensation of taste as they house the taste buds \cite{miller1990variations}, whereas {\em filiform papillae} that are devoid of taste buds are considered to be regulators of mechanoreception \cite{lauga2016sensing} for textural perception. Women are believed to have more fungiform papillae and are classed more frequently as supertasters \cite{bartoshuk1994ptc}. On the other hand, increased number of papillae have been found to be associated with enhanced fatty perception \cite{jilani2017association,zhou2021individual}. In addition to taste perception, papillae on the tongue are responsible for mechano-sensing. Mechano-sensing refers to our ability to sense the texture, friction, lubrication and touch on the tongue surface, and is carried out mainly by numerous filiform papillae that act as fine strain-amplified sensors on the tongue surface. These sensory functions are critical for manipulation and transport of food and liquids in the mouth\cite{lauga2016sensing,sarkar2019lubrication}. Such textural properties also influence our psychological reaction to food. For example, feelings such as satiety and therefore hunger are influenced by perception of friction and lubrication \cite{stribictcaia2020food,krop2019influence}. It has recently been shown that our preference for certain food such as chocolates is driven by surface lubrication that can be measured by artificial tongue-like surfaces~\cite{soltanahmadi2023insights}.  Besides food preferences, there is burgeoning interest in understanding the complex morphology of the tongue due to its involvement in various age-related oral conditions~\cite{tamura2012tongue,xu2019aging,hu2021dry}, mucosal degeneration and systemic diseases \cite{murphy2016dorsal,porter2017oral,huang2021sars,jin2020absence}. Certain medical conditions\cite{maeda2006dermoscopic} and inter-individual differences are known to be associated specifically with the morphology of the papillae and the tongue. Understanding the finer details in morphology, differences in papillae structures can thus lead to fabricating novel bio-inspired artificial surfaces in biomedical engineering, food engineering and therapeutics\cite{andablo20203d,arzt2021functional}.

\begin{figure}[ht!]
\centering
\includegraphics[scale=0.8]{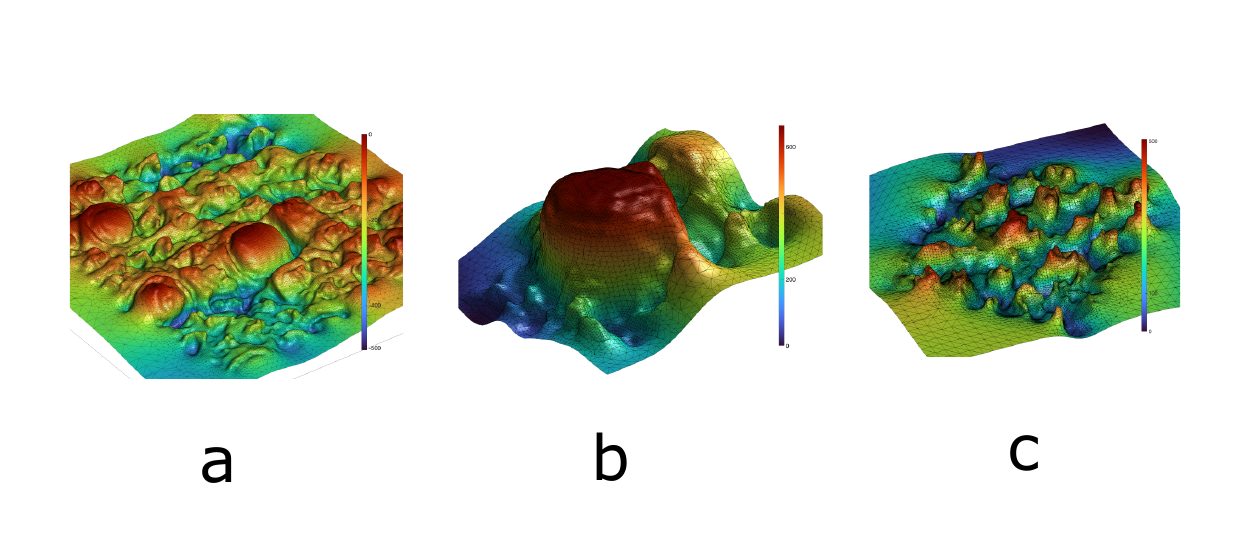}
\caption{\textbf{3D representation of a small portion of the dorsal part of the human tongue}. Plot (a) shows the 3D mesh of tongue surface obtained from masks taken on a real human tongue. The color bar shows the z-coordinate of the points on the surface representing the height. In plots (b) and (c) we see regions of the tongue with (b) Single Fungiform papilla and (c) Multiple filiform papillae. We note the distinctive shapes of papillae in plots (b) and (c), i.e, the dome-shaped Fungiform papilla in (b) and the crown-like shaped Filiform papillae in (c). Impressions of human tongue was collected at University of Leeds (Ethics DREC ref: 120318/AS/245, University of Leeds) \cite{andablo20203d}  from healthy adults (n = 15 subjects, 9 females, age 18–55 years)}
\label{fig:tongue_surface}
\end{figure}

The intricate geometry of the tongue at a microscopic scale can be appreciated in 3D scans (see Figure~\ref{fig:tongue_surface}). These images are obtained {\em via} surface reconstruction of 3D optical scans of a silicone-polymer mask of a human tongue. Fungiform papillae (Figure~\ref{fig:tongue_surface}(b)) are larger, sparsely distributed over the surface, and have a simple hemisphere-like shape. The average diameter of a fungiform papilla is about $878\mu\mathrm{m}$\cite{andablo20203d}, and they are clearly visible in larger images (Figure \ref{fig:tongue_surface}(a)). The filiform papillae show a more intricate crown shape (Figure~\ref{fig:tongue_surface}(c)). They are smaller (about $355\mu\mathrm{m}$ in diameter) and substantially more numerous. A square centimeter of human tongue surface is estimated to contain between $100$ and $200$ filiform papillae~\cite{andablo20203d}.  

Although there has been significant research on the importance of papillae density, our understanding of the papillae shapes and surface properties of the tongue suffers from the difficulty of extracting and analysing geometry of papillae at microscopic scales. Previous studies have thus focused on manually localising papillae from 2D images~\cite{nuessle2015denver}, primarily focusing on fungiform papillae \cite{cattaneo2020comparison}. Other works on biological surface data have used conformal geometry and computational topology at larger scales. Examples of such techniques include shape registration~\cite{hong2006conformal}, segmentation and topological data analysis\cite{amezquita2020shape,nicolau2011topology,oyama2019hepatic,krishnapriyan2021machine,saadat2021topological,khalil2022topological}. Machine learning has recently emerged as a powerful technique for diagnosis where large volumes of medical data or images are available~\cite{cai2020review}. These approaches have largely focused on computing global functions such as a medical diagnosis from an image. However, to date there is no machine learning model that has classified microscopic tongue papillae based on 3D tongue scans.

Herein, we present the first study of the 3D shapes of filiform and fungiform papillae in humans, with an emphasis on the variations in the microscopic geometry seen in Figure~\ref{fig:tongue_surface}. We develop a machine learning based framework applied to custom designed topological and geometric properties -- called {\em features} -- to understand one fundamental issue: {\em What separates one type of papillae from another?} We also ask the questions whether papillae are unique across and within individuals based on finer geometric details. Instead of applying machine learning as a black box application, we use statistics and explainable machine learning \cite{molnar2020interpretable,du2019techniques} to differentiate one type of papillae from another and identify the most distinctive features.

We follow the process of Topological Data analysis, where implicit shapes in data are extracted as topological {\em features} that form the basis of machine learning models. However, in addition to topological features, we also make use of geometric features computed from discrete curvatures to understand the uniqueness of tongue papillae. These features together are seen to have a high accuracy of $85\%$ in correctly identifying the papilla type (filiform or fungiform) in a small segment of a surface. As a result, we can now map the papillae arrangement for the first time -- including filiform papillae that are critical for developing biorelevant tribological surface and unravelling mechano-sensing -- as seen in Figure~\ref{fig:application_of_method}. 

Unprecedented analysis from our model reveals differences in papillae shapes across gender, age and individuals. We find that given a papilla, the age group and gender of the participant can be predicted to moderate accuracy, and even the exact individual from among $15$ participants can be identified with approximately $48\%$ accuracy, showing the first evidence for papillae to act as a unique identifier. This study demonstrating the uniqueness of papillae geometry at microscopic length scales using discrete differential geometry and computational topology stands to benefit future development of 3D tongue models for enabling rational food design diagnosis of oral medical conditions.

\section*{Results}

Our analytic framework processes the data, computes the features, and then applies machine learning driven analysis. We briefly explain the data processing and feature extraction. Then we proceed with a machine learning driven analysis of the feature set, prediction of gender, age and papillae type that reveals insights about papillae.

The data is obtained as 3D digital scans. The process starts with taking masks of the dorsal area of tongue of participants on silicone polymers. These masks are scanned using a 3D scanner, which yields a set of 3D points. These points are then passed through a surface reconstruction algorithm~\cite{kazhdan2013screened} implemented in Meshlab~\cite{cignoni2008meshlab}, which yields a mesh and a corresponding surface (see Figure \ref{fig:tongue_surface}). This process was developed by Andablo et al~\cite{andablo20203d}.

From this mesh data, we extract {\em segments} that are candidates for papillae. The extraction process is as follows. Around a point $P$ on the surface, select the set $B$ of points within a radius $r + \delta$, where $r=\max(r_{\mathrm{fungiform}}, r_{\mathrm{filiform}})$ $\mathrm{\mu m}$ and $\delta = 100\mathrm{\mu m}$, which we find to work well in practice. A plane fit to $B$ based on the RANSAC algorithm \cite{fischler1981random} represents our best approximation of the plane of the segment base. The {\em local maximum $m$} in the segment is defined  as the point furthest away from the plane. This point is assumed to  be the peak of a papilla, if present. Finally, we cut a region of radius $r$ around $m$ representing a candidate mesh for a papilla. Figure~\ref{fig:topological_information_varies}(a) and Figure~\ref{fig:topological_information_varies}(b) show such extracted segments for a fungiform and filiform papilla, while Figure~\ref{fig:topological_information_varies}(c) shows general surface area without any papilla. These three kinds of elements are the basis of our study.

A total of $2092$ segments extracted from scans of $15$ participants were labeled manually as Fungiform, Filiform or None. In the statistical workflow, a random subset of the segments (called the training set) is used to develop statistical models, while the remaining (the test set) -- whose labels are unknown to the model -- are used to test the accuracy of the models in a task of correctly predicting the label class (called {\em classification}). All accuracies reported in this paper are accuracy on the test set of unseen data. The analysis and machine learning are carried out on a large set of features (Table \ref{tab:feature-table-final}). In past work\cite{andablo20203d}, baseline features height and radii have been found to be distinctive between papillae types. Our more comprehensive segment dataset and computational models improve upon these baseline features to attain high accuracy automated detection of papillae type and other tasks.

\subsection*{Features and feature visualisation}

Features can be considered at different scales. At the global scale, a topological invariant of the entire papilla may be a distinctive feature. At the local scale of the neighborhood of a point on the surface, geometric properties -- in particular, curvature of points in the neighborhood -- best characterise the local shape of the surface. Local properties can be aggregated over the entire papilla to obtain a global feature. We describe below the significance of topological and geometric quantities in this context. 

\begin{figure}[ht!]
\centering
\includegraphics[scale=0.7]{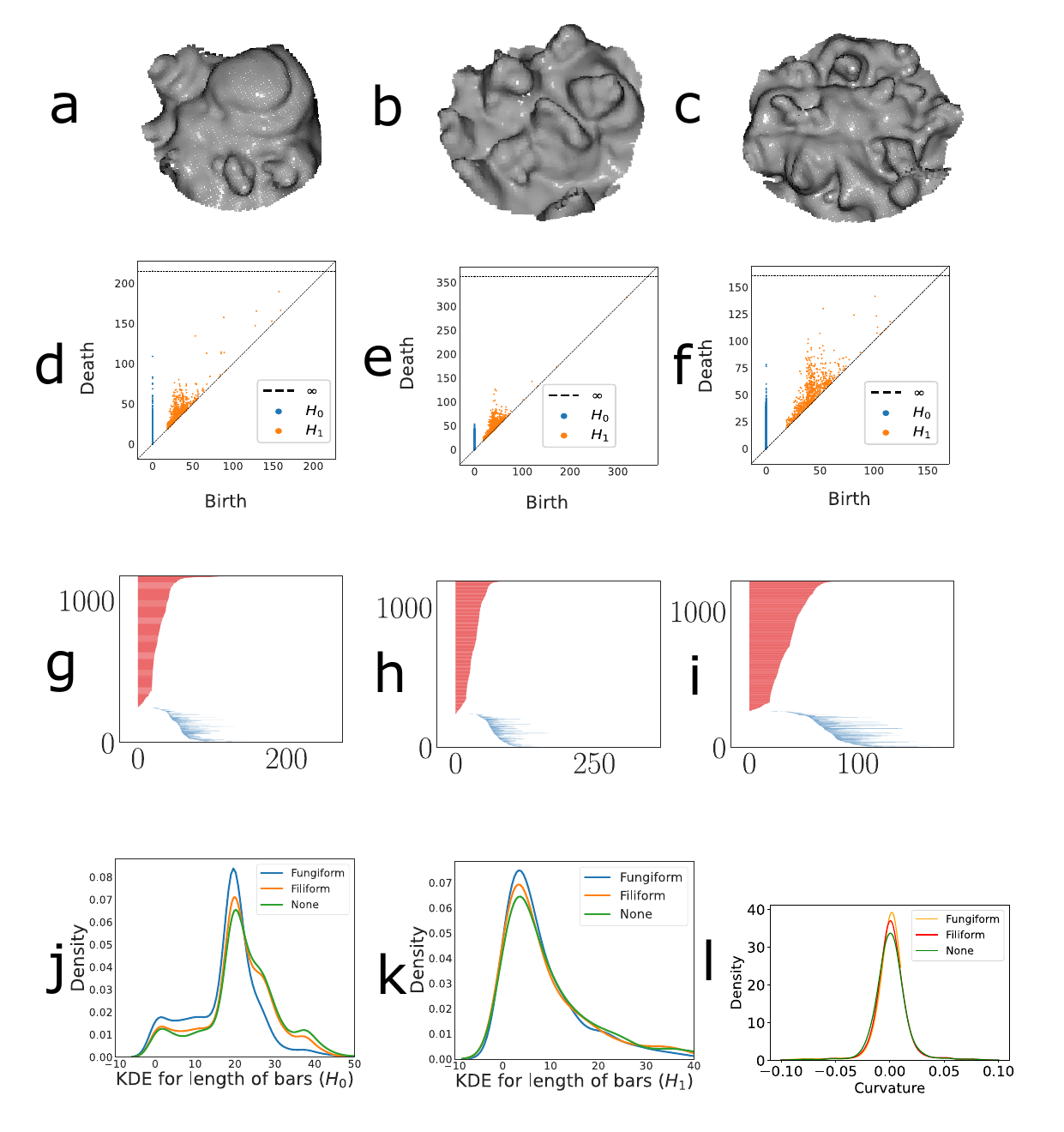}
\caption{\textbf{Papillae identification and topological feature characterization} Plots (a-c) show how the candidates for papillae from one Participant (Participant id 3) as meshes, using the library \texttt{open3d}. They are representatives from the 3 classes (a) Funigform, (b) Filiform and (c) None -- no papilla. Plots (d-f) show their respective topological representations of (a-c) in the form of persistent diagrams measuring two main topological features: $H_0$ -- the connected components and $H_1$ -- the equivalent loops. Plots (g-i) show the equivalent representation of the persistent diagram in the form of a barcode, where the bars in red correspond to the connected components and the bars in blue -- to the loops. Each bar represents a persistent generator, which is an interval where its left end point corresponds to the first filtration level where this topological feature appears, and its right end point is the filtration level where it disappears.
Plots (j) and (k) show the kernel density estimate (KDE) using Gaussian kernels -- plots representing the distribution of the lengths of bars from the barcode (for a,b and c.). Plot (l) reveals the curvature distribution across the different labels. }
\label{fig:topological_information_varies}
\end{figure}

\myparagraph{Topological features.} In this work, topological properties are computed via {\em persistent homology}. In this approach, each vertex (for us, a point on the reconstructed surface) is treated as the center of a growing ball, and the union of these balls is observed for changing topology. One way to interpret computational persistent homology is that it monitors topological features of different dimensions as they are born and die with the growth of the balls.
Connected components in $0$-dimension, loops in 1-dimension, and higher dimensional spheres in higher dimeensions. For a comprehensive introduction see the text by Edelsbrunner and Harer~\cite{edelsbrunner2022computational}. Figures \ref{fig:topological_information_varies}(d -- i) show the persistent topological components for the three types of segments, where the scale is measured in $\mathrm{\mu m}$. Figures \ref{fig:topological_information_varies}(d -- f) show the persistent diagram view, where each component manifests as a point indexed by its birth and death time. The difference in distribution of the points across plots suggests that there are variations in topological features for different segments. Figures \ref{fig:topological_information_varies}(g -- i) show an alternative view of the same data, called the barcode view -- where each bar shows the life duration of a topological component. From these sets of bars we can derive statistical features based on the distribution of bar lengths and more sophisticated methods. The feature which we have used in this work are based on persistent entropy, persistent images, persistence landscapes and amplitudes (please refer to the Methods section for detailed definition of each of the features and Table \ref{tab:feature-table-final}).

The distribution of bars at different lengths for $H_0$ (connected components) are shown in Figure \ref{fig:topological_information_varies}(j,k) as the kernel density estimates. Fungiform bar lengths in Plot~\ref{fig:topological_information_varies}(j) have higher density for shorter bars of length between $0$ and $10$ as compared to Filiform and None (around $0.01$), and then again in the mid range between $17$ and $25$, where all densities achieve their maximum. There are considerably fewer longer bars for Fungiform as compared to Filiform and None, which dominate the longer bar end of the spectrum. In plot (k) with densities $H_1$, we note that the density of short bars (lengths between 0 and 10) are higher for Fungifrom ($0.07$), followed by Filiform ($0.065$) and None ($0.06$). Thus there seems to be one predominant region of major difference, while $H_0$ shows greater variation across types.

\myparagraph{Geometric (curvature) features.} Curvature is locally defined at each point and is a complete descriptor of a surface. Positive curvature occurs where the surface matches a region of a sphere, for example at the top of a fungiform papilla. Sharp peaks are characterised by high positive curvature, while gentle tops, such as at the top of the fungiform papillae, have lower positive curvature. Negative curvatures are observed in saddle shaped neighborhoods, for example, around the base of papillae.

In digital discrete data, where manifolds are piecewise linear (triangulated) meshes, as in our case, curvature is computed at each vertex of the mesh as the angle deficit of the manifold (see Methods section for details). For our analysis, we compute curvatures on a sample of points in the segment. The geometric features of a segment include quantities such as the maximum and minimum of Gaussian curvatures, percentage of points with positive and negative Gaussian curvature, and other aggregated quantities (See Table \ref{tab:feature-table-final}).

The distribution of curvatures of the segments in Figure~\ref{fig:topological_information_varies}(a-c) are shown in Figure~\ref{fig:topological_information_varies}(l). For all types of papillae, most points are seen to be concentrated around small values of curvature close to zero. In particular, fungiform papillae have more points of near zero curvature, as can be expected from fungiforms having mostly flat or gently curving surfaces. In contrast, filliform and even generic surface areas are seen to have greater fraction of sharper curvature points. 

\myparagraph{Feature Visualisation.} The correlation matrix of features is shown in Figure \ref{fig:correlation_between_features} in the Supplementary material. This set of features were selected after removing features with correlation higher than $0.65$. The features remaining on this matrix show little correlation with each other, implying that they capture mutually distinct information, and thus they are informative in our analysis. Correlations by papillae type are shown in Figure \ref{fig:correlation_matrix_all_features}. 
PCA-based embedding of the data (Supplementary Figure \ref{fig:pca_plots_all_tasks}) shows overlap between classes. However, a non-linear method called Uniform Manifold Approximation and Projection for Dimension Reduction (UMAP)\cite{McInnes2018} does cluster the data in ways that show clear separation between classes (Figure \ref{fig:umap_plots_all_tasks}), implying implicit distinction between the classes. Next, we examine these features in order to quantify more closely their usefulness.

\begin{figure}[ht!]
\centering
\includegraphics[scale=0.7]{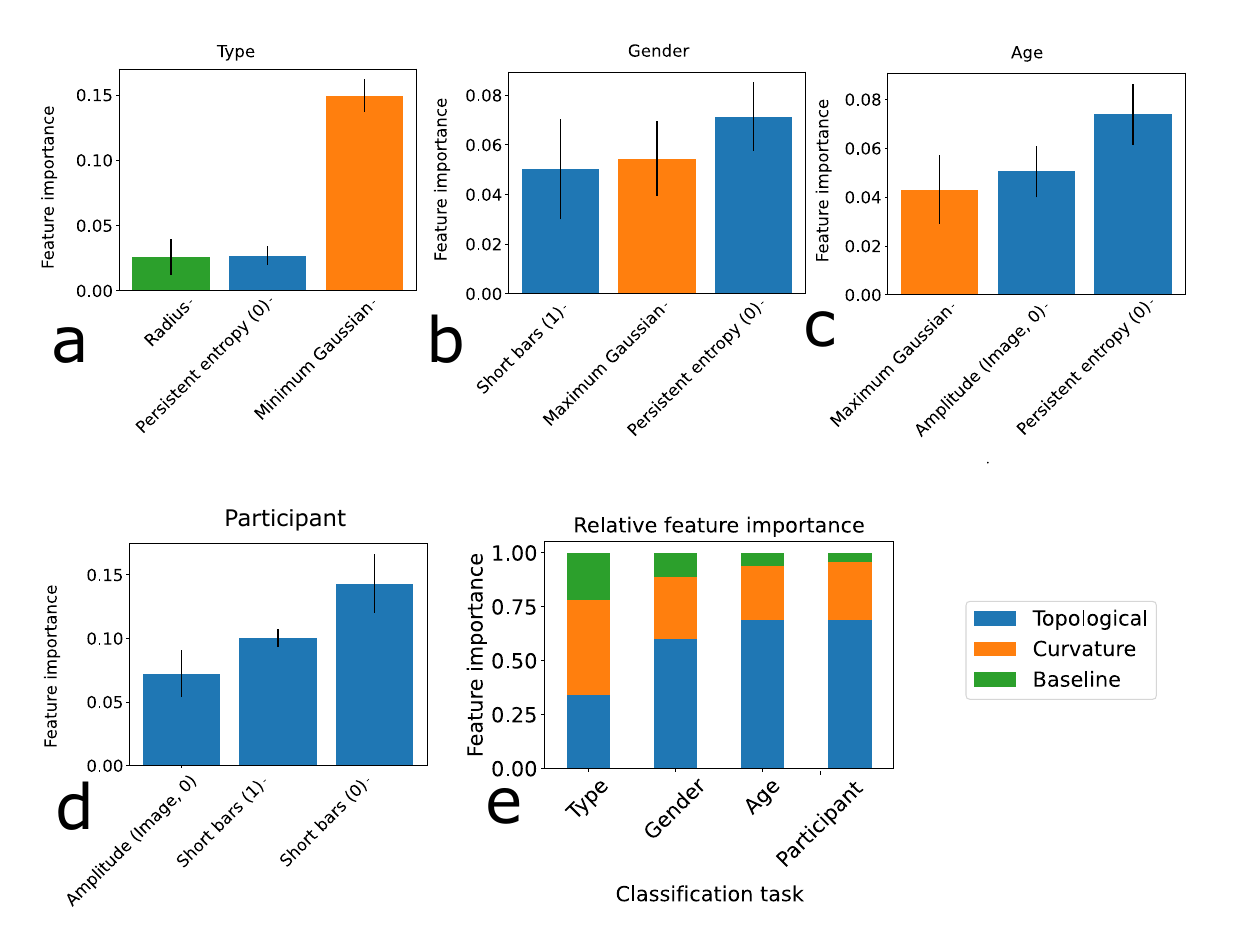}
\caption{\textbf{Feature importance across the classification tasks} The plots (a-d) represent the three most important features in the individual classification tasks. In particular, in plot (a) we see the papillae type task feature importance, in plot (b) Gender task features ordered by importance, in plot (c) the Age task features and in plot (d) the participant task features ordered by importance. The x-axis represents the accuracy drop when the feature of interest is permuted, and the black line represents the standard deviation over 30 runs. In plot (d) we see the relative importance of all features from each kind in each task. The curvature followed by topological features are the most important for papillae type classification; the topological are the most important for the Gender classification task; the topological are even more important for the Age classification task. We note the growing relative importance of topological features from $0.34$ to $0.69$ and the diminishing importance of the baseline features from $0.22$ to $0.04$, from left to right. The curvature features are the most important for the Type task with $0.44$ and maintain consistent medium importance across the Gender, Age and Participant prediction task with $0.28$, $0.25$ and $0.27$, respectively.}
\label{fig:feature_importance_between_features}
\end{figure}

\subsection*{Feature analysis and feature importance}

Various features may have different levels of importance in the distinction between papillae. The importance of a feature is a fundamental question in the field of explainable machine learning, and is usually  determined by its contribution to a classification model. It is a somewhat complex measure that is difficult to derive by looking at the feature in isolation. For our purposes, we use the technique called permutation feature importance \cite{breiman2001random}, and compute the contribution of these features to a class of standard classifiers called Kernel SVMs. The permutation feature importance method evaluates a feature $f$ by nullifying $f$ of the test data and observing the drop classification accuracy of the model. A large drop in accuracy implies $f$ is an important attribute for the classifier model. 
The effect of nullifying $f$ is achieved by permuting the values of $f$ among the test data points.

Figure~\ref{fig:feature_importance_between_features}, shows three most important features in determining each of the four labels of interest to us: the papillae type, the gender, the age and the participant id. The main observation here is that certain topological features are seen to be consistently important in these tasks (Figures~\ref{fig:feature_importance_between_features}(a-d)). Topological features overall are also found to contribute more to prediction accuracy than other features (Figure~\ref{fig:feature_importance_between_features}(e)).

\myparagraph{Type prediction features.} The KDE plots of the most important features for the papillae type classification task are presented in Figure \ref{fig:kde_type_feature_importance}, and the box plots and the aggregated distributions are shown in Supplementary Figure \ref{fig:important_papilla_classification_task}. The three distinctive features are seen to have very different distributions for the different types of segments, which explains their effectiveness in classification.

\begin{figure}[ht!]
\centering
\includegraphics[scale=0.7]{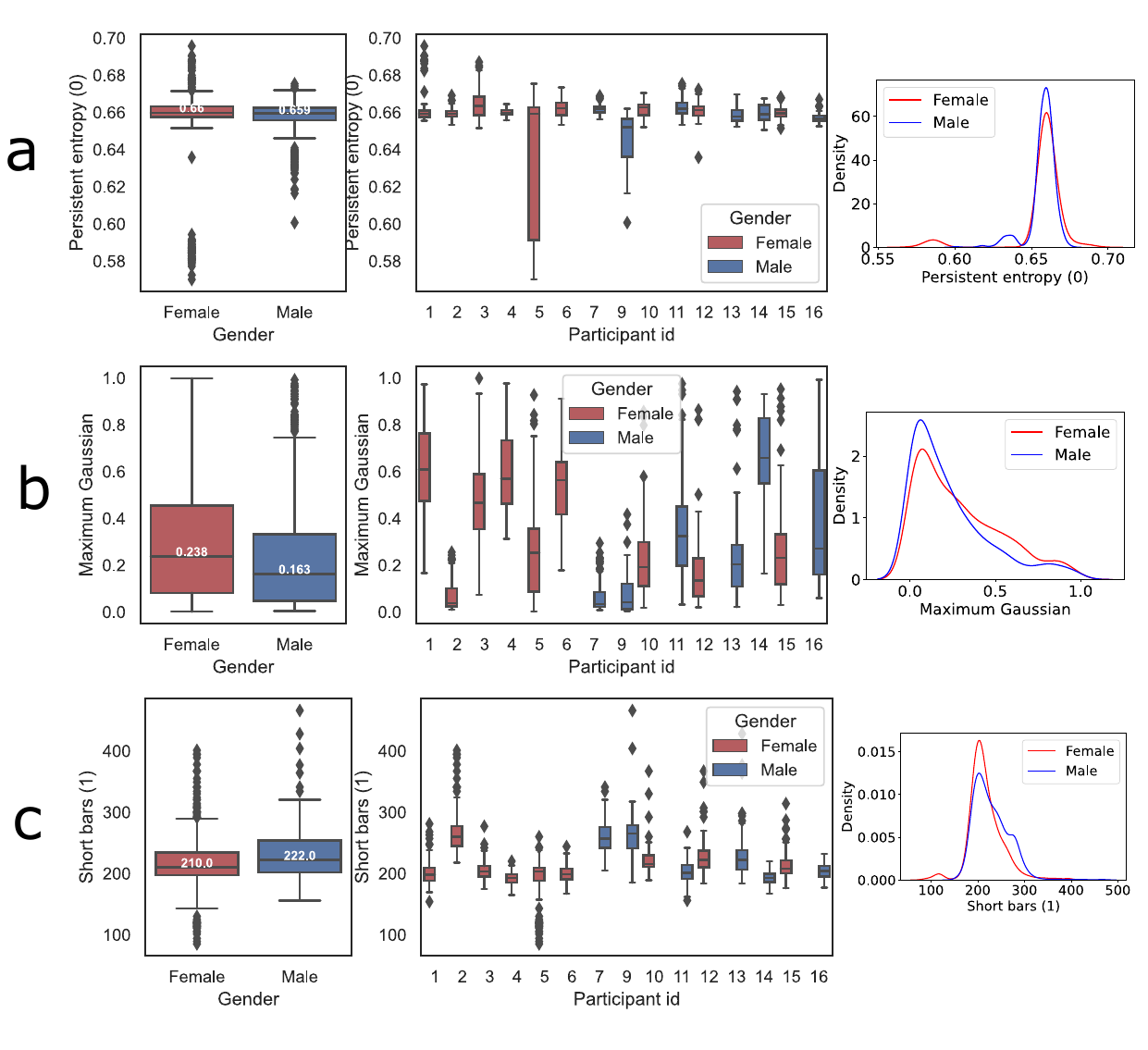}
\caption{\textbf{Important features for gender classification} The most important features for gender classification and its aggregate distribution. Both the aggregate and individual distributions show that the females have lower number of short bars than males.}
\label{fig:distribution_classification_task_gender_complete}
\end{figure}

\myparagraph{Gender prediction features.} We have two topological and one curvature feature at the top three for gender prediction task, whose box plots and aggregated distributions can be found in Figure \ref{fig:distribution_classification_task_gender_complete}. Persistent entropy (0) (Figure \ref{fig:distribution_classification_task_gender_complete}(a)), Maximum Gaussian curvature (Figure \ref{fig:distribution_classification_task_gender_complete}(b)) and Short bars (1) (Figure \ref{fig:distribution_classification_task_gender_complete}(c)) are all important features for determining gender. 
Figure \ref{fig:distribution_classification_task_gender_complete}(a), shows that the female participants tend to have a higher median value of the max Gaussian curvature (which holds for both Fungiform and Filiform) as compared to male participants, which could be linked to female papillae being `sharper', or `pointier'.

\begin{figure}[ht!]
\centering
\includegraphics[scale=0.6]{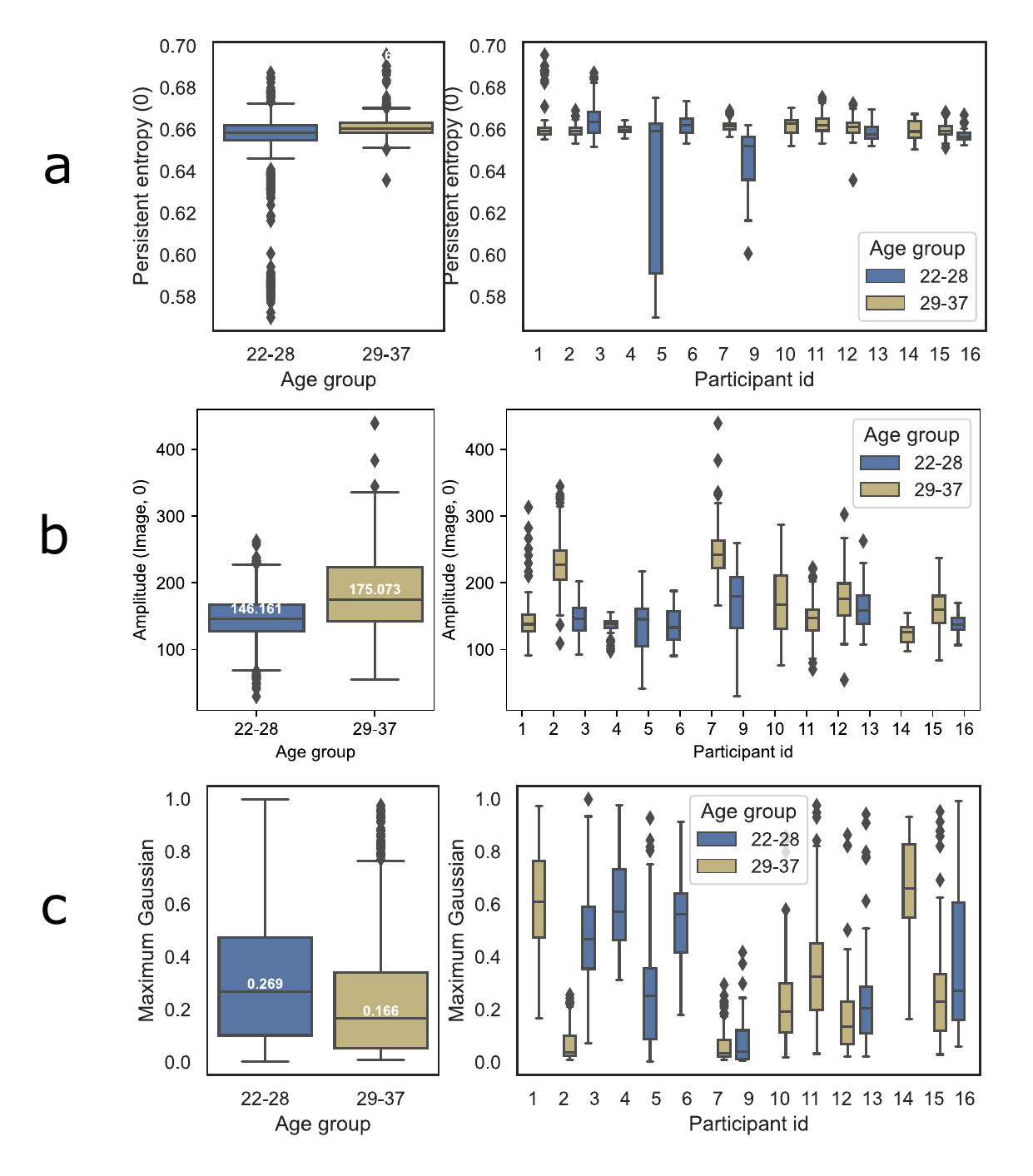}
\caption{\textbf{Important features for age classification} The most important features for age classification and its aggregate distribution.}
\label{fig:distribution_classification_task_age_complete}
\end{figure}

\myparagraph{Age prediction features.} Topological features also  dominate the age-prediction task, as seen in Figure \ref{fig:feature_importance_between_features}. The box plots and aggregated distributions of the top three features are presented in Figure \ref{fig:distribution_classification_task_age_complete} -- Persistent entropy (0)(Figure \ref{fig:distribution_classification_task_age_complete}(a)), Amplitude(Image,0) (Figure \ref{fig:distribution_classification_task_age_complete}(b)) and Maximum Gaussian (Figure \ref{fig:distribution_classification_task_age_complete}(c)) are the most important features for the age classification task. The baseline features (Height, Radius) are not amongst the most essential for this task, suggesting that their characteristics do not differ much for the two age groups in this study. An interesting observation is that height is more important than radius. The distributions can be seen in Figure \ref{fig:distribution_classification_task_age_complete}. Similar to the gender-prediction task, the Maximum Gaussian curvature feature (Figure \ref{fig:distribution_classification_task_age_complete}(b)) is one of the most important. The median for the younger age group is $0.269$ ($n = 840$) and for the older is $0.166$ ($n = 640$), implying some difference between the two groups, with the younger group having `pointier' papillae. This holds both for Fungiform and Filiform.

\subsection*{Predicting gender, age and participant from papillae structure}
Having understood the differences in papillae structure based on gender and age, we ask if one can easily predict gender, age and the participant given a papilla. Specifically, we ask if the papillae and the features identified above contain sufficient information to allow simple statistical methods to carry out accurate prediction. 

\subsubsection*{Gender prediction}

In this task we predict the biological gender of the participants. The classification performance is presented in Table \ref{tab:age-and-gender-results}. The models trained on topological features result in accuracy of $65\%$, outperforming the curvature features by $5\%$ and baseline features by $14\%$. Using all the features together marginally improves accuracy to $67\%$.

\subsubsection*{Age prediction}

The participants are split into two groups depending on their age. The cut-off is 29 to achieve a close to equal split. The classification statistics are shown in Table \ref{tab:age-and-gender-results}. The results follow similar pattern to the gender prediction task. The topological features on their own achieve classification accuracy of $0.73$, closely followed by curvature with $0.67$. The baseline features are behind by almost $0.10$, with a score of $0.58$. Combining the features once again improves accuracy to $0.75$. Results for Leave One Group Out test, where the age and gender of an unseen participant is predicted based on data from the others is shown in  Supplementary Table \ref{tab:age-and-gender-lopo-results}.

\begin{table}[htbp!]
\centering
\begin{tabular}{|l|l|l|l|}
\hline
\textbf{Model}        & \textbf{Balanced accuracy(Age)} & \textbf{Balanced accuracy(Gender)} & \textbf{Balanced accuracy(Participant)}\\
\hline
Baseline features               & $0.57 \pm 0.02$ & $0.52 \pm 0.03$  & $0.18 \pm 0.02$\\
\hline
Curvature features            & $0.66 \pm 0.02$  & $0.59 \pm 0.03$ & $0.22 \pm 0.02$ \\
\hline
Topological features           & $0.72 \pm 0.01$ & $0.65 \pm 0.02$ & $0.39 \pm 0.03$  \\
\hline
All Combined    & \textbf{0.74 $\pm$ 0.02} & \textbf{0.67 $\pm$ 0.02}  & \textbf{0.48 $\pm$ 0.02}\\
\hline
\end{tabular}
\caption{\textbf{Balanced accuracies for age, gender and participant prediction tasks} The topological features outperform the curvature and baseline features across all three tasks, and adding all features together does not improve the accuracy significantly for the age and gender tasks (only $0.02$ increase). However, this is not the case for the participant prediction task, where the performance improves with $0.09$. These results suggest that the topological information is a good indicator of age and gender. }
\label{tab:age-and-gender-results}
\end{table}

\subsubsection*{Participant identity prediction}
In this task we predict the participant from their papillae. The balanced accuracy of the topological features ($39\%$) are almost double  that of curvature features ($22\%$). This is illustrated by the most important features as well, as all three of them are topological. Unlike in the previous two tasks for gender and age, here combining all the features brings a significant improvement in the balanced accuracy score to $48\%$, suggesting that both the local and global information can contribute to predicting the identity of the participant. Note that while accuracies around $40\%$ to $50\%$ as seen here are not good on binary classification tasks, in this case the task is distinction among $15$ participants. A baseline rate of  random prediction in this case will produce an accuracy of only $6\%$. The features thus distinguish participants to a high degree of distinctiveness. 

\subsection*{Papillae detection and type classification.}
The final result is the accuracy of the classification task for 3-class classification (fungiform vs. filiform vs. none) based on the features (Table \ref{tab:feature-table-final}).  The classification statistics are shown in Table \ref{tab:three_class_balanced_accuracy}. The accuracy of the topological features is better than the baseline and the curvature features, and combining all features together provides the best accuracy. We achieve balanced accuracy of $0.72$ for the topological, $0.67$ for the curvature and $0.62$ for the baseline. Combining all the features increases the performance to $0.85$.

\begin{table}[htbp!]
\centering
\begin{tabular}{|l|l|l|l|l|l}
\hline
 & Bal. Acc (SVM) & Bal. Acc (LR) &  Bal. Acc (SVM-LOGO) & Bal. Acc (LR-LOGO) \\
\hline
Baseline (height, radius) &  0.62 $\pm$ 0.03 & 0.57 $\pm$ 0.03 & 0.59 $\pm$ 0.14 & 0.55 $\pm$ 0.11\\
\hline
Curvature (our method) & 0.67 $\pm$  0.03  & 0.60 $\pm$ 0.03 &  0.67 $\pm$  0.05  & 0.65 $\pm$ 0.03\\
\hline
Topological (our method) & 0.72 $\pm$ 0.03 & 0.67 $\pm$ 0.03 & 0.72 $\pm$ 0.08 & 0.69 $\pm$ 0.08\\
\hline
All Combined & \textbf{0.85 $\pm$ 0.02} & 0.80 $\pm$ 0.02  & \textbf{0.83 $\pm$ 0.05} & 0.80 $\pm$ 0.06\\
\hline
\end{tabular}
\caption{\label{tab:three_class_balanced_accuracy} Comparison of classification results for the classification task for 3-class classification (fungiform vs. filiform vs. none)} with random split and using Leave-One-Group-Out (LOGO), where the test data are taken from a single participant and training is carried out on samples from all other participants. The models used are Support vector machines (SVM) and Logistic regression (LR). The standard deviation for the baseline and topological features is larger for LOGO, suggesting that there is higher variation between participants for these feature sets. This is not the case for the curvature features, which appear to be more similar and stable across participants. However, when all the features are combined, the balanced accuracy is improved and the standard deviation is relatively low.
\end{table}

\subsubsection*{Application of classification model}
\begin{figure}[ht!]
\centering
\includegraphics[width=10cm]{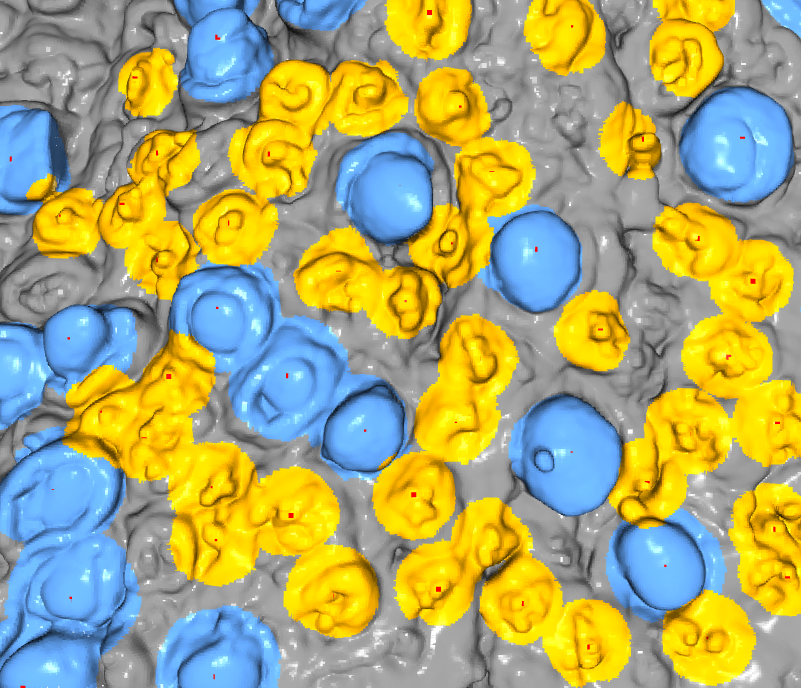}
\caption{\textbf{Automatic identification of tongue papillae} Illustration of the result of our tool for positioning papillae on the surface of the human tongue. Here our tool has detected the positions of fungiform (in blue) and filiform (in yellow) on the tongue surface. It has found 14 fungiform and 40 filiform papillae. As a red dot we see the centre of the papillae, which is determined as the local maxima for the structure with the highest distance from a fitted plane, using the RANSAC algorithm. }
\label{fig:application_of_method}
\end{figure}

The machine learning model developed can be used for accurate papillae detection and positioning on segments from a single person's tongue.  Figure~\ref{fig:application_of_method} shows the method accurately positions the fungiform form (in blue) and filiform (in yellow) on a tongue segment from one participant. This automated approach can thus efficiently and accurately construct maps or {\em tongue prints} from given tongue masks.

\section*{Discussion}
We have presented here the first study of the 3D shapes of human papillae based on high resolution scans. Our study is based on a novel framework combining geometry, topology and machine learning. Past research~\cite{sanyal2016tonguesim,valencia2016automatic} has focused on fungiform papillae in 2D images. In contrast, our microscale 3D reconstruction based approach can detect filliform papillae and non-papillated areas of the tongue, which are hard to distinguish with the naked eye and 2D images. Recent research has shown that the human perception of food is governed not only by the chemical sensation of {\em taste}, but also heavily by the mechanosensation, i.e. {\em texture} perceived by filliform papillae, for example, in the perception of soft textured delicacies such as chocolates~\cite{soltanahmadi2023insights}. Of more importance, the framework proposed here can be extended beyond the tongue papillae to the general study of shape and arrangement of microscale surface elements such as finger-like projections that are omnipresent in biology.

To capture the intricate biological shape information, we have developed a pool of geometric and topological features. While 3D geometric and topological transformations  have previously been used to process biological scan information~\cite{zhao2006lines,sundaram2008colon}, we employ a unique approach and treat them as statistical data that are fed to a machine learning system. In this approach, curvature statistics are used for aggregated local information, while persistent homology is used for global characteristics. Based on the subject of study, other features may be used. In our analysis topological features turn out to be more informative in prediction. Recent research\cite{bubenik2020persistent} has suggested that persistent homology can capture local shape information as well as global properties. Our results on tongue papillae are consistent with this idea. 

The analytics are based on machine learning models. The models themselves are built to predict the relevant variables of type, age, gender and participant, but our objective was to gain a better understanding of variations across classes and features. We thus used permutation feature importance to evaluate how each feature contributes to each model. From a pure accuracy point of view, large neural network models \cite{andreeva2020dr} trained on big datasets are considered the most successful current paradigm \cite{shahid2019applications}. However, our objective in this study  was to develop an interpretable framework for investigation of biological surface features, operating on relatively few samples from few participants. We have thus used simpler models that can be trained with smaller quantities of data. The accuracy of the results with simple models gives us confidence in our conclusion of feature importances and in the feasibility of highly accurate machine learning models in future research. 

The tasks for prediction of age group and gender suffer from the small number of participants. Machine learning models for these tasks achieve balanced accuracies of approximately $74\%$ and $67\%$ respectively. Note that for such binary prediction, a random prediction model achieves $50\%$ accuracy. The results suggest that geometric and topological features do vary to an extent across these variables, but more data will be needed to confirm the result and the nature of variation. The higher max Gaussian curvature appears as an important feature for female participants and the younger age group, suggesting more sharply curved or pointy shapes in these demographics. In past research, women and younger people have been noted to have higher density of fungiform papillae, which has been attributed to variations in taste perception, and women have been observed to be supertasters more frequently~\cite{bartoshuk1994ptc,fischer2013factors,zhang2009relationship}. The curvature variation implies a difference in papillae shapes that could be contributing to the sensory differences as well. Fungiform papillae density has been noted~\cite{karikkineth2021longitudinal} to drop above an age of $65$. In our study the participants were within the relatively young range of $22-37$. The shape features show some variations to reach a classification accuracy of $74\%$ between age groups $22-28$ and $29-37$. The Leave One Group Out test on age and gender (Supplementary Table \ref{tab:age-and-gender-lopo-results}) shows lower accuracy and greater variability. Certain individuals seem harder to model in this task. Further investigation with more participants will be required to gain greater insight into this issue.

The papillae type detection results are more accurate at $85\%$ and based on a large number of papillae, which gives us confidence that the model is truly accurate. To confirm that the models generalise to unseen participants, we carry out the Leave One Group Out test, and find that the accuracy holds up even on samples from a completely unseen participant, which confirms that the models can be used to classify and localise papillae on new tongue impressions. The papillae type model can thus be used to automatically identify filiform and fungiform papillae on scans of new tongue impressions. 

The individual participant model shows $48\%$ balanced accuracy and $51\%$ raw accuracy. This score is not impressive in a binary classification task, but our participant prediction task is a multi-class one, with $15$ possible classes. A papilla could have belonged to any one of the $15$ classes, and a random predictor would have an accuracy of only $6.66\%$. Considering the sample sizes from different participants, (Table \ref{tab:segments_per_participant}) a predictor that always predicts the largest class can achieve an accuracy of $11\%$. In comparison, the model achieves between $4$ to $8$ times the accuracy of these baselines based on the distinctiveness in the data of a single papilla. This distinctiveness may have multiple contributing factors -- these can be true inter-individual variations as well as variations in experimental conditions in collecting the masks.  The exact cause of this difference will require further study. Note that while the age, gender and participant identification tasks suggest unique individual characteristics, the success of the type identification task suggest a complementary conclusion of significant similarity within types and across individuals. Larger studies can potentially address some of these issues using larger models and more complex features, such as persistent homology of curvature functions.

The framework and discriminative models presented here enable deeper study of the papillae structure and their variations and arrangements. The model for localising and classifying papillae (as seen in Figure~\ref{fig:application_of_method}) enables the study of the overall tongue surface, or {\em tongue prints}. Such arrangements of papillae are known to influence the surface properties of the tongue and its perception abilities~\cite{andablo20203d}. Our data and past research have shown that the distribution of papillae vary across individuals. A detailed study of this variation across various demographic parameters could reveal insights into preferences, cultures and medical conditions. Arrangements identified by our models could be used to build generative models that can fuel such insights and can create more realistic surfaces for use in food engineering and development of oral diagnostics. Ultimately, this study offers a new dimension showing papillae as an unique identifier for the first time in the literature which needs further validation using this developed method for a larger dataset of participants. 

\section*{Methods}

\subsection*{Data collection}
\subsubsection*{Collection of Human Tongue Silicone Impressions.}
The data in this study has been obtained from 3D optical scans of masks of real human tongues from 15 healthy participants performed using an Alicona InfiniteFocus (IF), details of the data collection has been described in a previous publication~\cite{andablo20203d}. Negative impressions of the upper surface of the tongue were collected from ($n = 15$ subjects, the mean age in years is $29.1$, SD=$3.7$), 6 male and 9 female. More detailed information can be found in Table \ref{tab:demographics_table_gender_age}.

\subsubsection*{Experimental protocol.}
The study adhered to all relevant guidelines and regulations. Signed informed consent was obtained from all participants before undertaking the experimental protocol. The ethics declaration is included at the end of this section. 

\subsection*{Dataset generation for papillae}

Each participant's point cloud was split into two smaller parts of approximate size $13\mathrm{mm}$ by $9\mathrm{mm}$ in order to reduce the size of the point cloud. On each part, The Screened Poisson surface reconstruction\cite{kazhdan2013screened} in Meshlab\cite{cignoni2008meshlab} is applied. Then, a number of circular segments of radius $r + \delta$, where $r$ is set to match $max(r_{fungiform}, r_{filiform})$\textmu m and $\delta = 100$, were extracted according to our algorithm for extracting candidates for papillae locations described below. Based on previous work~\cite{andablo20203d} and our experiments in detecting papillae, we find that $r_{fungiform} = 439, r_{filiform} = 177.5$ work well for automated detection. These segments have been manually labelled into one of three classes: fungiform papillae, filiform papillae or None (neither a fungiform nor a filiform). The final dataset consists of 414 fungiform, 1489 filiform and 190 None, resulting in 2092 tongue segments in total. The number of segments per participants can be found in Table \ref{tab:segments_per_participant}.

\subsubsection*{Finding Candidates for papillae locations}
The pipeline for segment extraction works as follows. First, we pick a random point $P$ on the surface. Then, we select a radius $r + \delta$ of points around $P$, where we set $r$ to match $max(r_{fungiform}, r_{filiform})$\textmu m. Note that the distance from $P$ is computed as the $3D$ Euclidean distance in the ambient space. If the set of points contain disconnected components, then components not containing $P$ are discareded from the computation. We set $\delta = 100$ to fully cover any papilla in the region. After that, we fit a plane based on the RANSAC algorithm \cite{fischler1981random} and identify the point $M$ furthest away from the plane, which will be a {\em local maxima}. We identify it to be the centre of the segment. Finally, we cut a region of radius $r$ around $m$ as a candidate segment. This process is applied repeatedly to identify multiple papilla segments. In future iterations, any maximum within a previously processed segment is ignored. Number of iterations and samples in our experiments were limited by the need for manual labelling. In applying our model for mapping papillae (e.g. as in Figure~\ref{fig:application_of_method}) the process can be continued until no new papillae is found.

\subsection*{UMAP for visualisation}

UMAP \cite{mcinnes2018umap} represents data by fitting it to non-linear manifolds and thus can capture complex information. We have used the supervised version of the method for visualisation. The supervised method explicitly tries to separate known classes by embedding their connectivity graphs. We use it to test presence of distinction between classes.

\subsection*{Baseline, curvature and topological features}
Three sets of features are extracted from each of the selected segments -- baseline, curvature and topological.

\subsubsection*{Baseline features}
We use geometric measurements for baseline feature identification, which comprises of two quantitative shape characteristics of the papillae: height and radius. From the data presented in Table 1 in \cite{andablo20203d}, based on Tukey's test for statistical significance of the means and standard deviation, the diameter (and the radius, respectively) and height are different between fungiform and filiform. Therefore, they can serve as features for distinguishing between the three classes. 

We note that defining the height and radius automatically is a challenging task due to the irregular nature of these structures and to our knowledge no unambiguous definitions exist in the literature to date to identify these features accurately. Human participants do this manually by observing the continuity of the papillae from the base to the tip.

We compute height and radius as follows. The point $m$, identified as the local maximum for the segment, as the centre of the structure. Then we define the radius $r$ as the radius value of the sphere, centered at $M$, which contains $90\%$ of the points in the segment. We compute this iteratively, by first guessing the value of the radius $i$ as a small value ($100\mu$m), and count the number of points in the neighbourhood of radius $i$ (we use KDTree with FLANN \cite{muja2009flann} for nearest neighbor search). We then increase $i$ by $10$, until the number of points contained in the neighbourhood exceeds $90\%$ of all points. The value of $i$ at the stopping condition is our candidate for radius value, $r$.
The computation of the height, $h$, is dependent on the value of $r$. It works as follows: we first cut a region around the centre of radius $r$, we then fit a plane using the RANSAC algorithm \cite{fischler1981random} and find the maximum distance from the plane to the local maximum point $M$. This value is our height value, $h$. All the computations have been performed using the Python libraries \texttt{open3d} and \texttt{numpy}. The algorithm is mimicking the manual procedure which a tongue expert would use to compute these values. An illustration of the procedure can be found in Figure S8.

\subsubsection*{Curvature features}

For each $x \in H$, where $H$ is the surface generated by the Poisson surface reconstruction process, we compute the discrete curvature as defined by Meyer et al.~\cite{meyer2003discrete}. The definition in the discrete case on the triangular mesh is via the vertex's angular deficit $k_H(v_i) = 2\pi - \sum_{j \in N(i)}^{} \theta_{ij}$, where $N(i)$ are the triangles incident on vertex $i$ and $\theta_{ij}$ is the angle at vertex $i$ in triangle $j$. The way that Gaussian and mean curvatures are computed uses averaging Voronoi cells and the mixed Finite-Element/Finite-Volume method~\cite{meyer2003discrete}. We use the existing implementation from the Python version of Meshlab, called \texttt{pymeshlab} \cite{muntoni2021pymeshlab}.

We use the maximum and minimum of the Gaussian and mean curvature as features, the ratio of positively curved points to the number of all points in the mesh ($k_{positiveratio}$), and we introduced a new feature called curvature ratio ($k_{ratio}$). Let $x$ be the number of points of positive curvature, and $y$ be the number of points of negative curvature. Therefore, we define the curvature ratio $k_{ratio}$ to be $k_{ratio}= \frac{y}{x}$ if $y \leq x$  and $k_{ratio}=\frac{x}{y}$ if $x \leq y$. The signs of the mean and the Gaussian curvature provide plenty of information about the local behavior of the surface \cite{colombo20063d}. We computed the discrete Gaussian and mean curvature for all meshes and calculated the number of vertices of positive and negative curvature (after the Poisson surface reconstruction filter). The ratio of positively curved points to the number of all points in the mesh is defined as $k_{positiveratio}=  \frac{x}{x + y}$. The full list and intuitive interpretations are provided in the Supplementary material, Table S4.

\subsubsection*{Topological features}

We subsample the 3D point clouds to 1000 points each and compute the Vietoris-Rips complex, using the Euclidean distance as a filtration. Persistent homology \cite{edelsbrunner2022computational} of the 3D point cloud was computed using the \texttt{giotto-tda} library~\cite{tauzin2021giotto} and \texttt{ripser}~\cite{bauer2021ripser}. We then generate 12 features which are one number summary of the diagram, providing different topological information. For more details on persistent homology, please refer to the Supplementary material.

\textit{Short bars} are the number of intervals of length between 0 and 10. We compute them both in homology dimension 0 and 1. This features has been found to capture the local geometry of an object \cite{bubenik2020persistent}

\textit{Persistent entropy}\cite{chintakunta2015entropy,atienza2020stability} is the measure of the entropy of the points in a persistent diagram. Concretely, let $D = \{(b_i, d_i)\}_{i \in I}$ be a persistent diagram with non-infinite death times, i.e., $d_i < \infty$. Then, the persistence entropy of $D$ is defined as $P_E(D) = \sum_{i \in I}p_{i}\log(p_{i})$, where $p_i = \frac{(d_i - b_i)}{L_D}$ and $L_D = \sum_{i \in I}(d_i - b_i)$. We compute persistent entropy in dimension 0 and 1, and denote it by Persistent entropy (0) and Persistent entropy (1). 

\textit{Persistence landscapes}: Given a persistent diagram $D = \{(b_i, d_i)\}_{i \in I}$, its persistence landscape is the set $\{\lambda_k \}_{k \in \mathbb{N}}$ of functions $\lambda_{k}(t):\mathbb{R} \rightarrow [0, \infty]$, where $\lambda_{k}(t)$ is the $k$-th largest value of the set $\{g_{(b_i,d_i)}(x)\}^{n}_{i=1}$, where $g_{(b,d)} = 0$ if $x \notin (b, d)$; $g_{(b,d)} = x - b$ if $x \in (b, \frac{b+d}{2})$ and $g_{(b,d)}=-x + d$ if $x \in (\frac{b+d}{2}, d)$. The parameter $k$ is called a layer. In this work we consider the case when $k = 1$.

\textit{Persistence image}: diagrams are converted to sums of Dirac deltas. The convolution with Gaussian kernel is performed, where the computation is done over a grid with rectangular shape. The locations of the points are evenly sampled from the values of the filtration, turning it into a raster image, which is then flattened into a vector. 

\textit{Amplitude} can be defined as the distance from the persistent diagram to the empty diagram, which contains only the diagonal points. Here we use 2 kernels (persistence landscapes \cite{bubenik2017persistence} and persistence image \cite{adams2017persistence}) and the amplitude of the kernel is computed using the $L2$ norm, and 2 metrics (Wasserstein and Bottleneck). For the computation, we use the default parameters in \texttt{giotto-tda}. 

We here denote Persistence image amplitude by Amplitude (Image, 0) Amplitude (Image, 1) for the computation of the amplitude with the persistent image kernel (which is the is the $L2$ norm of that vector) in homology dimension 0 and 1, respectively. Similarly, Amplitude (Landscape, 0) and Amplitude (Landscape, 1) is the Persistence Landscape amplitude in homology dimension 0 and 1. 

\textit{The Wasserstein amplitude} of order $p$ is the $Lp$ norm of the vector of point distances to the diagonal, which is $A_{w} = \frac{\sqrt2}{2}(\sum_{i \in I}(d_i - b_i)^{p})^{\frac{1}{p}}.$ Here we use $p = 2$. Similarly, the \textit{Bottleneck amplitude}, $A_B$, is defined by letting $p$ to $\infty$ in the definition of the Wasserstein amplitude. In other words, it is a fraction of the longest bar $A_{B} = \frac{\sqrt2}{2}\sup_{i \in I}(d_i - b_i)$. We denote them by Amplitude (Wasserstein, 0), Amplitude (Wasserstein, 1) and Amplitude (Bottleneck, 0), Amplitude (Bottleneck, 1) respectively, corresponding to the different homology dimensions.

\subsection*{Machine learning and statistics}

\myparagraph{Classification models.} The experiments use classes of simple models -- Support vector machines (SVMs) and Logistic regression models. The implementations from \texttt{scikit-learn} \cite{pedregosa2011scikit} were used without modification and with the default hyperparameters. The SVMs were used with a radial basis kernel (RBF). Details of these techniques can be found in any introductory book on machine learning. We use $20\%$ of the data for testing and the other $80\%$ for training using a random split. The procedure is repeated $50$ times. 

\myparagraph{Performance Metrics for machine learning}. Accuracy represents the proportion of correct predictions made by the model out of the total number of predictions. To adjust for the varying number of samples across classes, we compute the balanced accuracy. It calculates the average of the correct classification proportions for both positive and negative observations.

\myparagraph{Feature Importance.} The plots are based on classification by the best balanced accuracy split of the data, and 30 permutations of the features for that split. The black line represents the standard deviation of the feature importance over the 30 runs.


\subsection*{Ethics declarations}

Signed informed consent was obtained from all participants before undertaking the experimental protocol. Ethical approval for this study was granted by the University of Leeds ethics committee DREC ref: 120318/AS/245, as well as the University of Edinburgh (Reference number 2019/71645).

\section*{Data availability}
The datasets generated and analysed during the current study are not publicly available but are available on reasonable request to authors.





\section*{Acknowledgements}


RA is supported by the United Kingdom Research and Innovation (grant EP/S02431X/1), UKRI Centre for Doctoral Training in Biomedical AI at the University of Edinburgh, School of Informatics. This project has received funding from the European Research Council (ERC) under the European Union’s Horizon 2020 research and innovation program (Grant Agreement No. 757993). Dr. Efren Andablo-Reyes (School of Food Science and Nutrition, University of Leeds) and Dr. Paul Hydes (School of Dentistry, Faculty of Environment, University of Leeds are kindly acknowledged for collection of tongue masks and primary data development. The authors would like to thank Camille Hammersley (School of Mechanical Engineering, University of Leeds) for her technical support in using the Alicona InfiniteFocus instrument for 3D optical scanning of the tongue masks.

\section*{Author contributions statement}

A.S. has collected the data, R.A. and R.S. conceived the experiment(s),  R.A. conducted the experiment(s), R.A analysed the results. R.A. wrote the first draft. All authors discussed the results, commented on, revised and reviewed the manuscript. 

\section*{Additional information}

The authors declare that there are no competing interests.




\bibliography{sample}

\clearpage
\renewcommand{\theequation}{S\arabic{equation}}
\renewcommand{\thetable}{S\arabic{table}}
\renewcommand{\thefigure}{S\arabic{figure}}
\setcounter{equation}{0}
\setcounter{table}{0}
\setcounter{figure}{0}

\section*{Supplementary material}
\begin{figure}[ht!]
\centering
\includegraphics[scale=0.7]{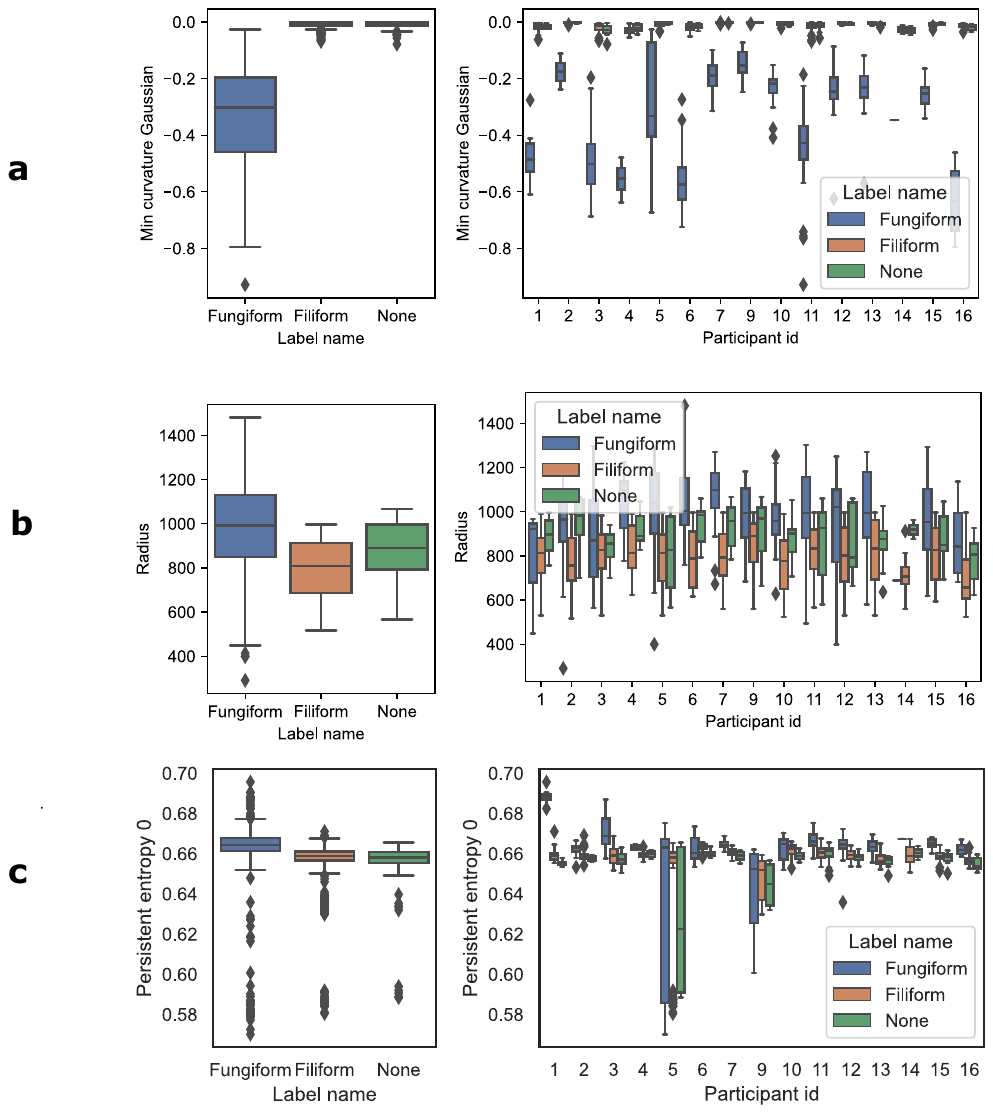}
\caption{\textbf{The most important features for papillae type classification} (a-c) are the features with the highest importance for the papillae type classification task. (a) and (c) are topological, while (b) is curvature. }
\label{fig:important_papilla_classification_task}
\end{figure}

\begin{table}[ht!]
\centering
\begin{tabular}{|l|l|L|L|}
\hline
\textbf{Feature type} & \textbf{Feature name}       & \textbf{Description} & \textbf{Intuitive interpretation} \\
\hline
\multirow{2}{*}{Baseline features}               
& Radius   & The radius of the papillae & The horizontal distance between the projection of the top and the border \\ \cline{2-4}
              & Height   & The height of the papillae & The vertical distance between the top of the papillae and the base \\ 
\hline
\multirow{8}{*}{Curvature features}           & Minimum Gaussian   & The minimum value of the Gaussian curvature &  How pointy downwards an object is\\
\cline{2-4}
            & Maximum Gaussian   & The maximum value of the Gaussian curvature  & How pointy the object is at its maximum\\
\cline{2-4}
            & Ratio Gaussian   &  The ratio with (+) $k_{Gaussian}$ over (-) $k_{Gaussian}$, if $\#$ of (+) $\leq$ $\#$ (-); and the other way round  &  Approximately measuring the papilla curves towards the normal\\ 
\cline{2-4}
            & Ratio mean   & The ratio with (+) $k_{mean}$ over (-) $k_{mean}$, if $\#$ of (+)  $\leq$ $\#$ of (-); and the other way round & Approximately measuring how positively curved the papilla is\\
\cline{2-4}
            & Positive Gaussian   & The percentage of points with positive Gaussian curvature  & A measure of how positively curved the papilla is, what percentage is dome-like\\
\cline{2-4}
            & Positive mean   & The percentage of points with positive mean curvature  & The percentage of points with positive mean curvature\\
\hline
\multirow{8}{*}{Topological features}           & Persistent entropy (0)   & The Shannon entropy of the barcode in $H_0$ & Measuring how different the lengths of the bars are in $H_0$ \\
\cline{2-4}
    & Short bars (0) and (1)  & The number of short bars in $H_0$ and $H_1$, respectively  & The number of least persistent connected components in $H_0$ and loops in $H_1$ with relatively short life span \\
\cline{2-4}
 & Amplitude (Bottleneck) (0) and (1)   & The distance between the persistence diagram and the empty diagram in the Bottleneck metric  & Approx. measurement of the length of the longest bar, or the life span of the most persistent feature \\
\cline{2-4}
 & Amplitude (Persistence image)   & The distance between the persistence image and the empty diagram  & Quantity measuring the topology of the object and how much it differs from a flat surface \\
\hline
\end{tabular}
\caption{Description of non-correlated baseline, curvature and topological features.}
\label{tab:feature-table-final}

\end{table}

\begin{figure}[ht!]
\centering
\includegraphics[scale=0.7]{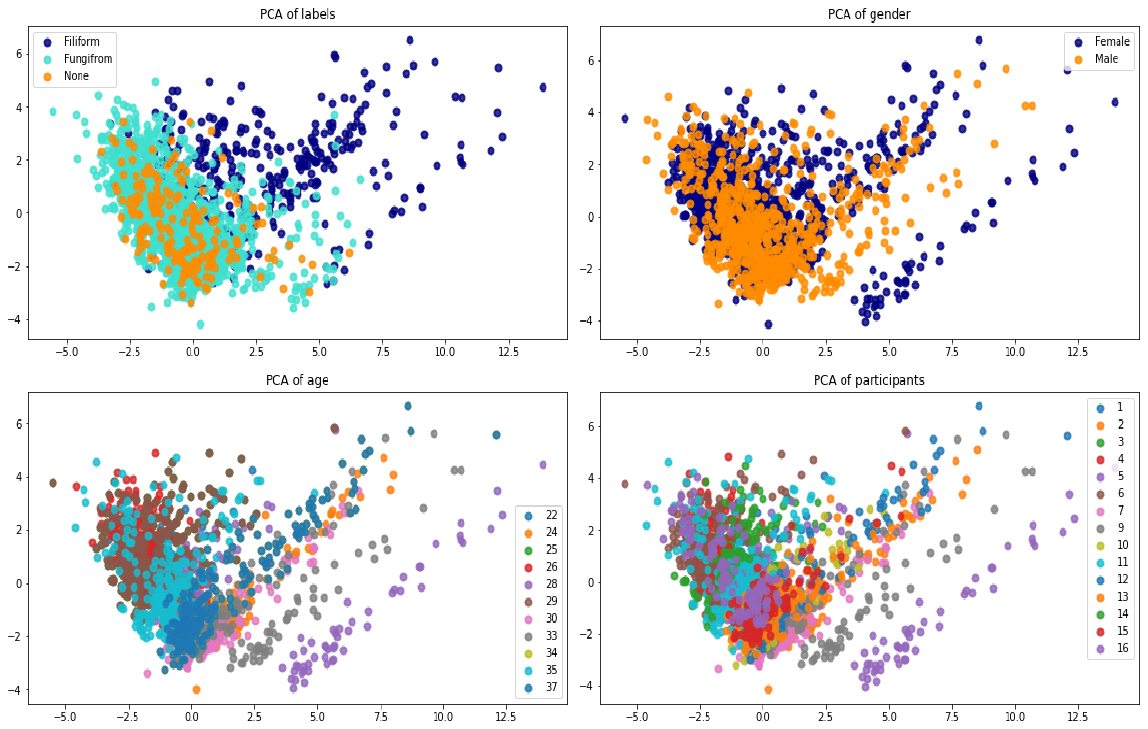}
\caption{Here we see PCA plots for all the features. In the first row on the left, we see the labels for the 3 classes. The second plot in the first rows shows the gender. In the third plot we see the different ages of the participants. In the fourth plot we see the PCA plot for the participants. We can see that participant 9 and participant 5 are clustered very nicely. The remaining participant's features also exhibit some sort of clustering though not as clear as the others, which is an interesting observation.}
\label{fig:pca_plots_all_tasks}
\end{figure}

\begin{figure}[ht!]
\centering
\includegraphics[scale=0.7]{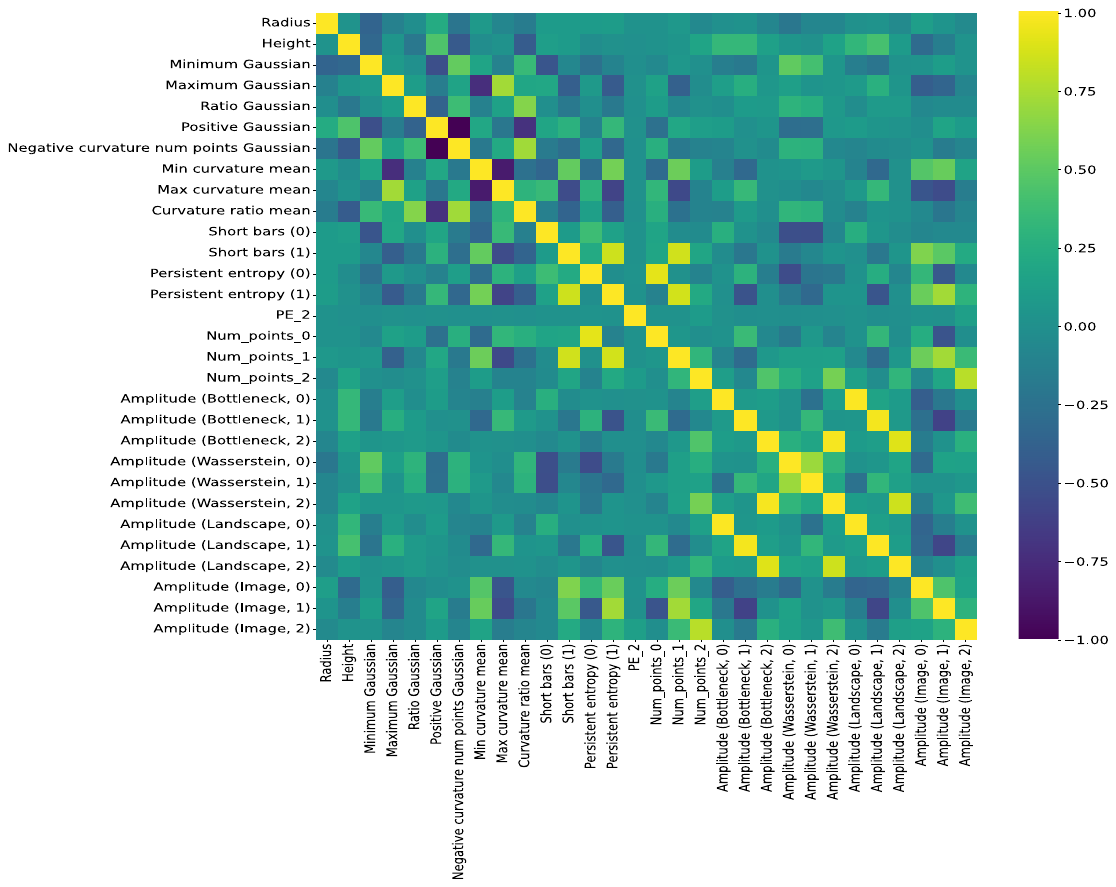}
\caption{Correlation matrix for all features}
\label{fig:correlation_matrix_all_features}
\end{figure}

\begin{figure}[ht!]
\centering
\includegraphics[scale=0.7]{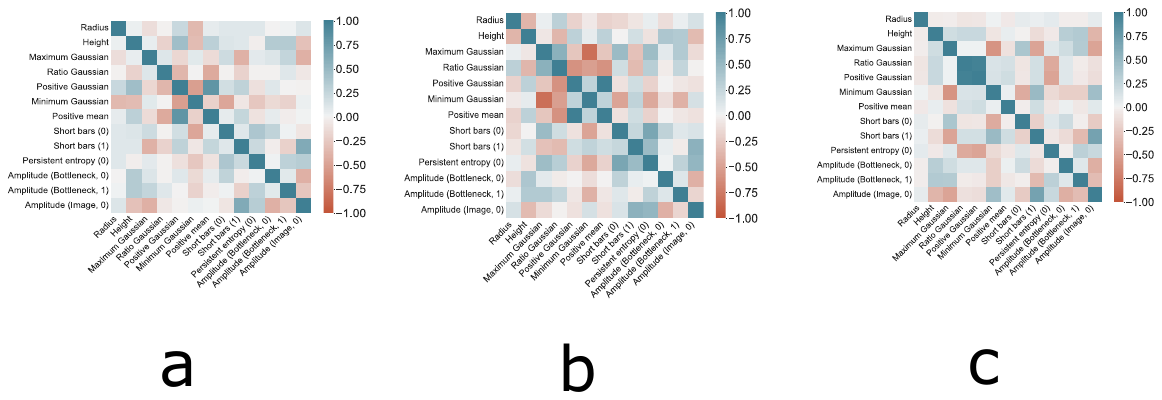}
\caption{\textbf{Correlation matrices} (a) is the correlation matrix for all papillae types, (b) is the correlation matrix for fungiform and (c) is the correlation matrix for filiform. We filtered down to this set of features by removing other features that had high correlation (more than $0.65$). The features remaining on this matrix show little correlation, implying that they capture different properties of papillae shape. Thus they are the basis of our analysis. However, once we filter by papillae type, some higher correlation is visible. For example, when we look at the correlations for the features of fungiform only, there is almost perfect correlation between Positive mean and Positive Gaussian. This was hidden in the overall correlation matrix.}
\label{fig:correlation_between_features}
\end{figure}

\begin{figure}[ht!]
\includegraphics[width=\linewidth,scale=0.9]{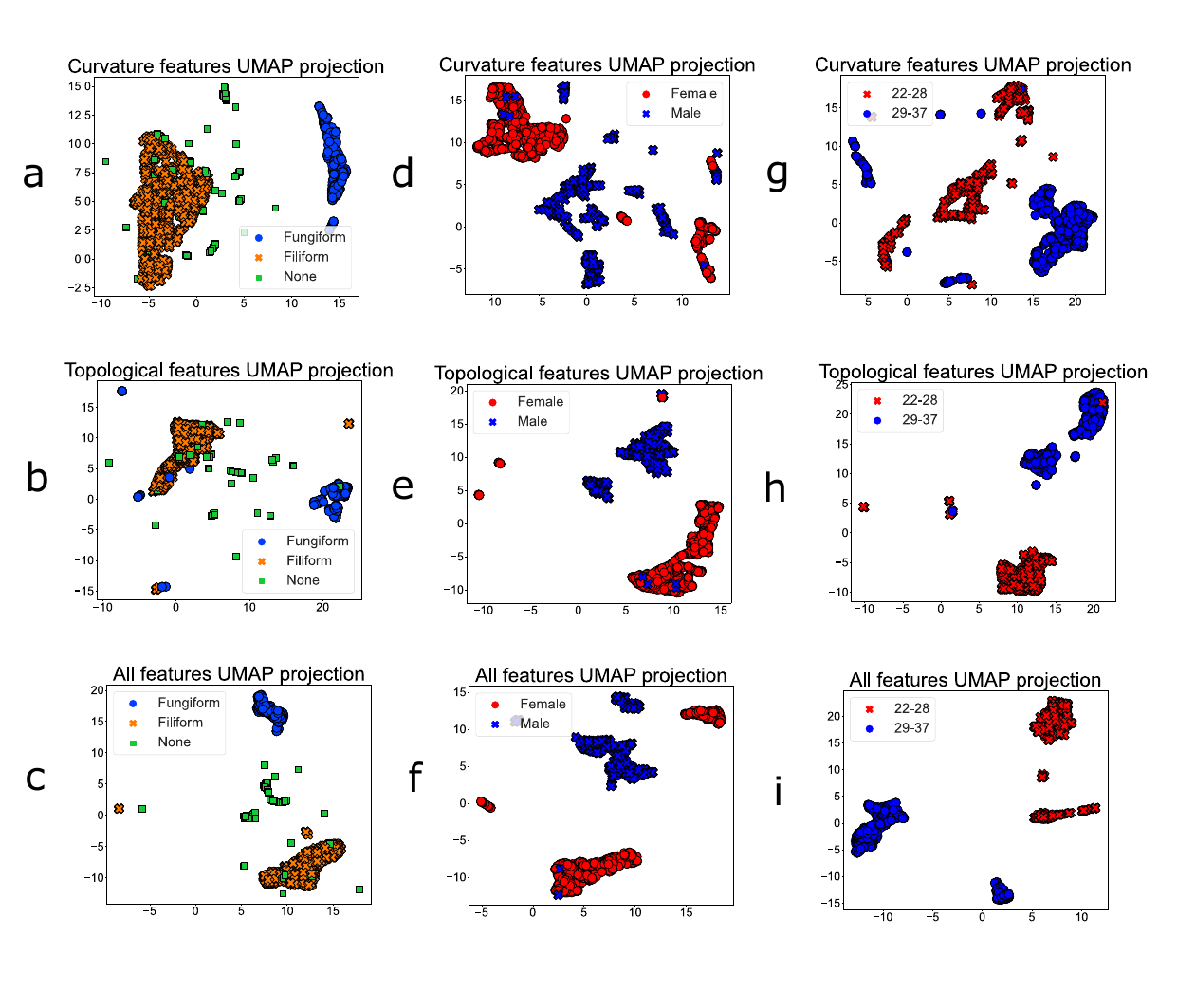}
\caption{UMAP plots for three classification tasks. (a-c) are for papillae type, (d-f) are for gender and (g-i) are for the age task. The clusters are more well defined when we look from the top to the bottom (from curvature to all features), which is also demonstrated by the performance at the classification tasks -- the topological features outperform the curvature, and the combination of all features together achieves the best performance.}
\label{fig:umap_plots_all_tasks}
\end{figure}

\begin{figure}[ht!]
\centering
\includegraphics[scale=0.7]{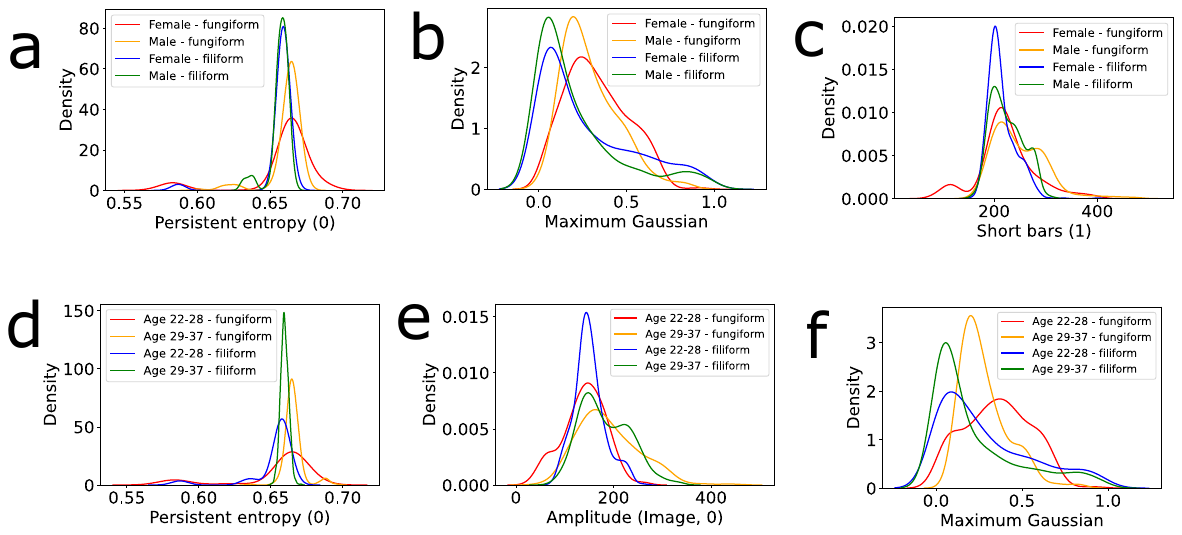}
\caption{\textbf{Comparison between the papillae type for the most important gender and age features} .}
\label{fig:gender_age_kde_plots}
\end{figure}

\begin{figure}[ht!]
\centering
\includegraphics[scale=0.7]{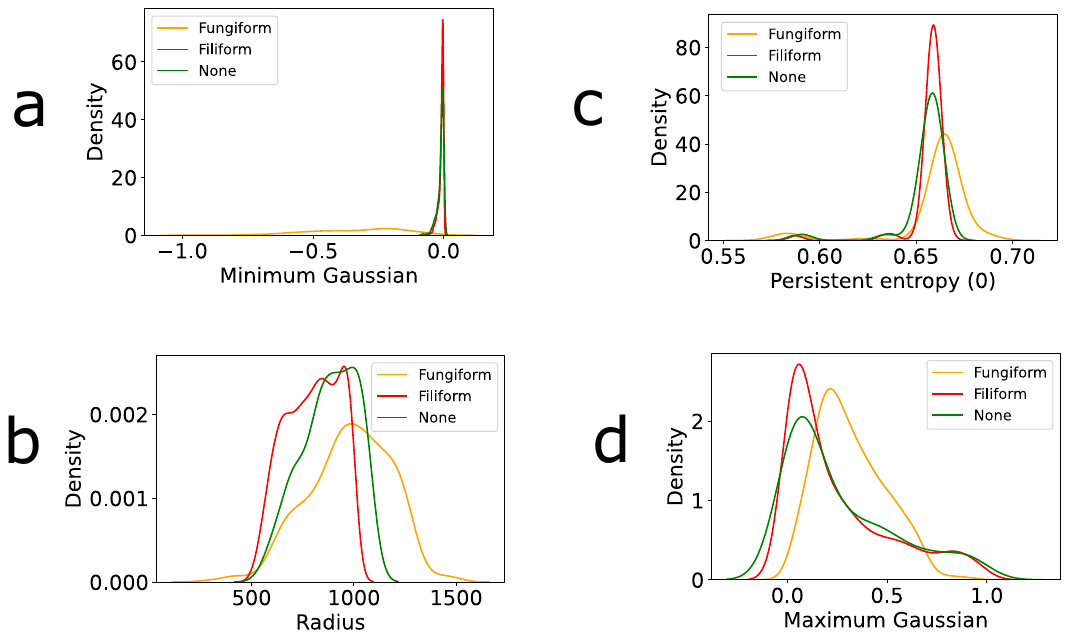}
\caption{\textbf{KDE plots of most important features for type classification task}. In plot (a), the distribution of the Minimum Gaussian for Fungiform follows a very different pattern from Filiform and None -- it is mostly flat and evenly distributed on the interval $(-1,0)$, while Filiform and None are densely concentrated around $0$. In plot (b), the distribution of Filiform and None are very similar, with Fungiform having higher value of Radius, which is as expected from previous work \cite{andablo20203d}. In plot(c), the papillae Types Filiform and None follow a similar pattern  to (a). Fungiform has higher value of Persistent entropy (0). In plot (d), even though Maximum Gaussian is not amongst the top three most important features, the value for Maximum Gaussian is higher for Filiform and None compared to Fungiform. This is as expected due to the sharper shape of filiform with more pronounced drop and steeper sides.}
\label{fig:kde_type_feature_importance}
\end{figure}

\begin{table}[ht]
\centering
\begin{tabular}{|l|l|}
\hline
Participant id  & Segments \\
\hline
1 & 173\\
\hline
2 & 163\\
\hline
3 & 168 \\
\hline
4 & 149 \\
\hline
5 & 199\\
\hline
6 & 138\\
\hline
7 & 139\\
\hline
9 & 115\\
\hline
10 & 77\\
\hline
11 & 240\\
\hline
12 & 132\\
\hline
13 & 106\\
\hline
14 & 51\\
\hline
15 & 121\\
\hline
16 & 121 \\
\hline
\end{tabular}
\caption{\label{tab:segments_per_participant}Number of segments per participant.}
\end{table}

\begin{table}[ht!]
\centering
\begin{tabular}{|l|l|}
\hline
Gender & Age (mean;SD)\\
\hline
 Female & 29.5 (4.5) \\
 \hline
 Male & 28.3 (3.9) \\
\hline
\end{tabular}
\caption{\label{tab:demographics_table_gender_age} Demographics of the participants.}
\end{table}

\begin{table}[htbp!]
\centering
\begin{tabular}{|l|l|l|l|l|}
\hline
\textbf{Model}        & \textbf{Balanced acc(Gender, all)} & \textbf{Balanced acc(Gender, lim)} & \textbf{Balanced acc(Age)} & \textbf{Balanced acc(Age,lim)}\\
\hline
Baseline features               & $0.45 \pm 0.12$ & $0.52 \pm 0.06$  & $0.52 \pm 0.11$ & $0.52 \pm 0.11$\\
\hline
Curvature features            & $0.44 \pm 0.25$  & $0.67 \pm 0.15$ & $0.57 \pm 0.30$ & $0.59 \pm 0.28$ \\
\hline
Topological features           & $0.45 \pm 0.29$ & \textbf{0.67 $\pm$ 0.11} & $0.57 \pm 0.30$ & \textbf{0.61 $\pm$ 0.27} \\
\hline
All Combined    & \textbf{0.42 $\pm$ 0.28} & $0.65 \pm 0.14$  & $0.50 \pm 0.24$ & $0.53 \pm 0.21$\\
\hline
\end{tabular}
\caption{Balanced accuracies for age and gender tasks. The performance when we use Leave-one-group-out approach. The results are worse than before due to a small number of participants having too low accuracies. It will be a question for future work to investigate this. The results on the right are when these small number of participants have been removed. It can be that they are outliers and their topological features do. However, it is difficult to make conclusions given the small sample size. It could be that some people are topological outliers, and their features are not similar at all to the other people in the same age category}
\label{tab:age-and-gender-lopo-results}
\end{table}

\begin{figure}[ht!]
\centering
\includegraphics[scale=0.7]{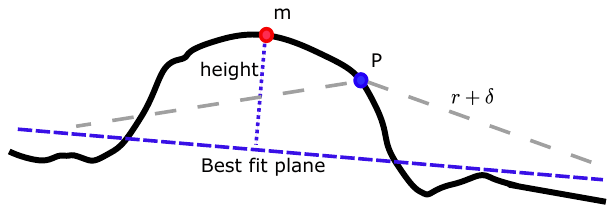}
\caption{\textbf{Schematic figure: Profile view of processing a segment with a fungiform papilla.} From an arbitrary point $P$, all mesh vertices within a radius $r+\delta$ are taken. Then a Best fit Plane is found using the RANSAC algorithm. The candidate point for the peak of a papilla (if present) is found as $m$ -- the point furthest from the plane. This distance is taken to be the height, and $m$ is assumed to be the centre of the papilla.}
\label{fig:schematic_image}
\end{figure}

\clearpage









\end{document}